\pdfoutput=1

\documentclass[11pt]{article}

\usepackage{EMNLP2022}

\usepackage{times}
\usepackage{latexsym}

\usepackage[T1]{fontenc}

\usepackage[utf8]{inputenc}

\usepackage{microtype}

\usepackage{inconsolata}

\usepackage{textcomp}
\usepackage{stfloats}
\usepackage{url}
\usepackage{verbatim}
\usepackage{graphicx}

\usepackage{algorithm}
\usepackage{algorithmic}
\usepackage{amssymb}
\usepackage{helvet}  
\usepackage{courier}  
\usepackage{newfloat}
\usepackage{listings}
\usepackage{multirow}
\usepackage{amsmath}
\usepackage{booktabs}
\usepackage{pifont}
\usepackage{color}

\newcommand{{\model}}{KP-PLM}

\title{Knowledge Prompting in 
Pre-trained Language Model for \\ Natural Language Understanding}

\author{Jianing Wang$^1$, Wenkang Huang$^2$, Qiuhui Shi$^2$, Hongbin Wang$^2$, \\
{\bf Minghui Qiu$^3$, Xiang Li$^1$\thanks{\ \ \ Correspondence to Xiang Li.} , Ming Gao$^{1,4}$} \\
$^1$ School of Data Science and Engineering, East China Normal University, Shanghai, China \\
$^2$ Ant Group, Hangzhou, China\ \ \ 
$^3$ Alibaba Group, Hangzhou, China \\
$^{4}$ KLATASDS-MOE, School of Statistics, East China Normal University, Shanghai, China \\
\texttt{lygwjn@gmail.com, wenkang.hwk@alibaba-inc.com}\\
\texttt{\{qiuhui.sqh,hongbin.whb\}@antgroup.com}\\
\texttt{minghui.qmh@alibaba-inc.com}\\
\texttt{\{xiangli,mgao\}@dase.ecnu.edu.cn}
}
\begin{document}
\maketitle

\begin{abstract}
Knowledge-enhanced Pre-trained Language Model (PLM)
has recently received significant attention, 
which aims to incorporate factual knowledge into PLMs. 
However,
most existing methods modify the internal structures of fixed types of PLMs by stacking complicated modules, 
and introduce redundant and irrelevant factual knowledge from knowledge bases (KBs).
In this paper,
to address these problems,
we introduce a seminal 
knowledge prompting paradigm
and further propose a knowledge-prompting-based PLM framework \model.
This framework can be flexibly
combined with existing mainstream PLMs.
Specifically, we first construct a knowledge sub-graph from KBs for each context.
Then we design multiple continuous prompts rules and transform the knowledge sub-graph into natural language prompts. 
To further leverage the factual knowledge from these prompts,
we propose two novel knowledge-aware self-supervised tasks 
including prompt relevance inspection and masked prompt modeling. Extensive experiments on multiple natural language understanding (NLU) tasks show the superiority of {\model} over other state-of-the-art methods
in both full-resource and low-resource settings~\footnote{All the codes and datasets have been released to~\url{https://github.com/wjn1996/KP-PLM}}.
\end{abstract}

\section{Introduction}

Pre-trained Language Models (PLMs) have become the dominant infrastructure for a majority of downstream natural language understanding (NLU) tasks,
which generally adopt a two-stage training strategy, 
i.e., \emph{pre-training} and \emph{fine-tuning}.
Recent
notable PLMs include BERT~\cite{Devlin2019BERT}, RoBERTa~\cite{Liu2019RoBERTa}, ALBERT~\cite{Lan2020ALBERT}, XLNet~\cite{Yang2019XLNet}, GPT-2~\cite{radford2019language} and DeBERTa~\cite{He2021Deberta}. 
Although many existing models learn useful inherent \emph{linguistic knowledge} from large-scale corpora, 
it is hard for them to understand the explicit \emph{factual knowledge}~\cite{Robert2019Barack, Sun2020CoLAKE}.

\begin{figure}
\centering
\includegraphics[width=\linewidth]{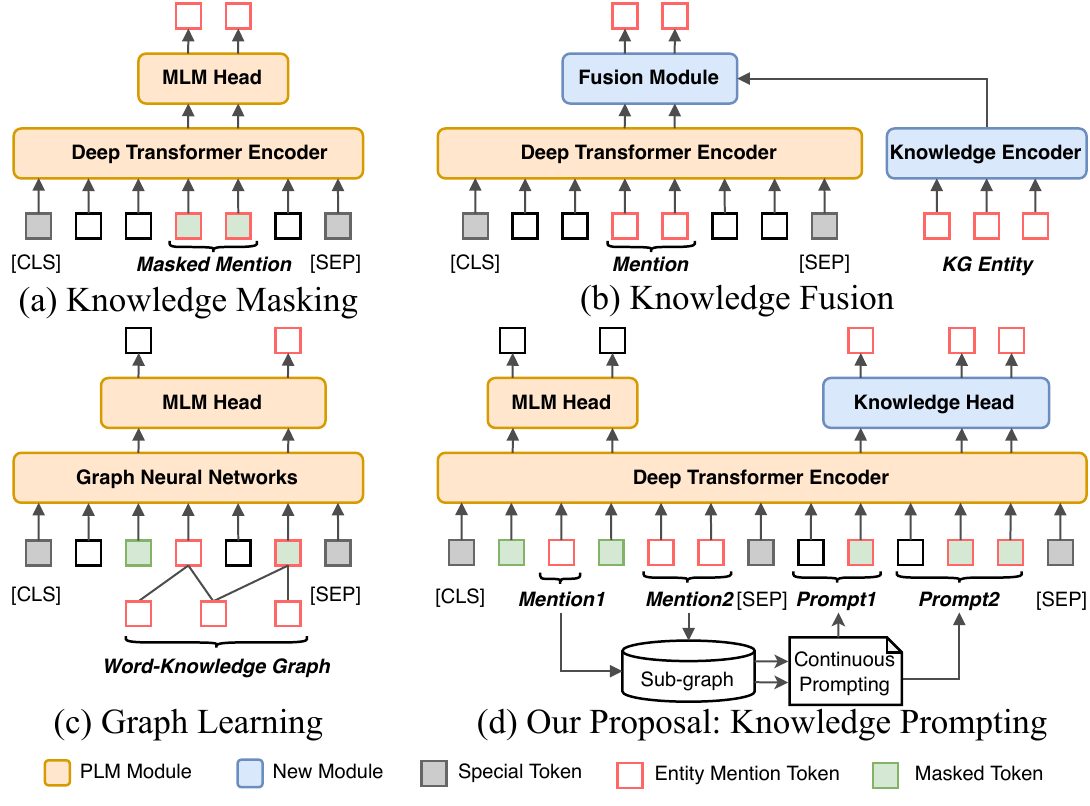}
\caption{
The
comparison between {knowledge prompting} with other paradigms. 
(Best viewed in color.)
}
\label{fig:overview}
\end{figure}

To further leverage factual knowledge,
a branch of knowledge-enhanced methods~\cite{Zhang2019ERNIE, Sun2020CoLAKE,He2021KLMo,Wang2021KEPLER,Wang2021KAdapter, Zhang2021DKPLM, Arora2022Metadata} have been proposed for PLMs
to capture rich semantic knowledge from knowledge bases (KBs). 
Figure~\ref{fig:overview}
summarizes the paradigms of
existing approaches, 
which can be mainly divided into three categories.
First, 
\emph{knowledge-masking-based methods} 
perform alignment,
mask mentions
and learn explicit knowledge of entities 
based on masked language modeling (MLM). 
Second, 
\emph{knowledge-fusion-based methods} separately learn embeddings of entities in sentences and that of nodes in KBs,
which are further aggregated.
Third, 
\emph{graph-learning-based methods}
construct a graph based on 
contextualized sentences and KBs,
and
adopt
a graph-based encoder to learn entity embeddings, such as 
graph neural networks~\cite{Kipf2017Semi}.
Despite the success,
there could still be 
two problems in these methods. 
On the one hand, 
some approaches modify internal structures of existing PLMs 
by stacking complicated modules, 
which adversely affects the computational cost of the model.
Further,
these methods are generally based on fixed types of PLMs, 
which leads to 
the inflexibility 
and restricts their wide applicability~\cite{He2021KLMo,Zhang2021DKPLM}.
On the other hand, 
some methods 
introduce 
redundant and irrelevant knowledge from KBs, which are the knowledge noises and could degrade the model performance~\cite{Peters2019Knowledge}:
1) Redundant Knowledge. Previous methods~\cite{Sun2020CoLAKE, Liu2020KBERT, Sun2019ERNIE, Wang2021KEPLER} inject the corresponding triples or pre-trained knowledge embeddings into each entity in the context. However, the entity may appear multiple times in one sentence, these methods could introduce duplicate information, which can be viewed as redundant knowledge. 2) Irrelevant Knowledge. Some entities or the corresponding sub-graphs are irrelevant with the whole sentence semantics, which are useless and have less contributions to the performance improvement.

Recently, the paradigm of prompt-based fine-tuning (i.e., prompt-tuning) has been proposed to 
explore the inherent knowledge of PLMs by 
introducing a task-specific template with the \texttt{[MASK]} token~\cite{Schick2021Exploiting}\footnote{For example, in sentiment analysis, a prompt template~(e.g., ``It was \texttt{[MASK]}.'') is added to the review text (e.g., ``The film is very attractive.''). We can obtain the result tokens of masked position for label prediction (e.g., ``great'' for the positive label and ``boring'' for the negative label).}.
Inspired by 
prompt-tuning,
in this paper,
we introduce a novel \emph{knowledge prompting} paradigm for knowledge-enhanced PLMs
and 
propose an effective \textbf{K}nowledge-\textbf{P}rompting-based \textbf{PLM} framework, namely,~{\model}. 
As shown in Figure~\ref{fig:overview}(d), 
we aim to construct the related knowledge sub-graph for each sentence.
To alleviate introducing too much knowledge noises, we restrict all the relation paths start from the topic entity (the entry of each sentence) and end at the tail entity which is mentioned in this sentence.
To bridge the structure gap between the context and KB, we can transform each relation path into a natural language prompt,
and then concatenate these prompts with the original sentence to form a unified input sequence,
which can be flexibly fed into any 
PLMs without changing their internal structures~\cite{Brown2020Language, Schick2021Exploiting}. 
Further, 
to leverage the factual knowledge in the prompts,
we propose two novel knowledge-aware self-supervised learning tasks: \emph{Prompt Relevance Inspection} (PRI) and \emph{Masked Prompt Modeling} (MPM).
Specifically, PRI encourages the PLM to learn the semantic relevance of multiple knowledge prompts, while MPM aims to predict the masked entity in a prompt.
We conduct extensive experiments to verify the effectiveness of~{\model} on multiple NLU tasks. Our results show that \model\ can be flexibly integrated with mainstream PLMs and the integrated model can outperform strong baselines in both full-resource and low-resource settings.
We next summarize our main contributions as follows:






\begin{itemize}
    \item We propose a novel knowledge prompting paradigm to incorporate factual knowledge into PLMs. 
    
    \item We present the~{\model} framework that can be flexibly combined with mainstream PLMs. 
    
    \item 
    We design two knowledge-aware self-supervised tasks 
    to learn factual knowledge from prompts.
    
    \item 
    We conduct extensive experiments to show the effectiveness of \model\ in both
    full-resource and low-resource scenarios.
\end{itemize}


\section{Related Work}


\noindent\textbf{Knowledge-enhanced PLMs.}
To further inject factual knowledge into PLMs, recent knowledge-enhanced PLMs have been proposed to incorporate factual knowledge in the pre-training stage~\cite{Sun2019ERNIE, Xiong2020Pretrained, Liu2020KBERT, Sun2020CoLAKE, Zhang2019ERNIE, He2021KLMo, Zhang2021DKPLM, Su2021CokeBERT, Arora2022Metadata, Ye2022ontology, Yu2022JAKET, Yu2022Dict, Chen2022DictBERT, Liu2021KG}.
For example, ERNIE-Baidu~\cite{Sun2019ERNIE} introduces two novel phrase-level masking and entity-level masking strategies 
to learn explicit semantic knowledge. 
K-BERT~\cite{Liu2020KBERT} and CoLAKE~\cite{Sun2020CoLAKE} utilize contextualized sentences and KBs to construct a graph, 
based on which graph-based learning methods are employed to capture semantic information.
In addition, ERNIE-THU~\cite{Zhang2019ERNIE}, KEPLER~\cite{Wang2021KEPLER} and KLMo~\cite{He2021KLMo} integrate the pre-trained knowledge base embeddings into PLMs through attentive fusion modules to improve the contextual representations.
Different from these methods, 
we propose a novel knowledge prompting paradigm to enhance PLMs,
which learns factual knowledge from prompts.

\noindent\textbf{Prompting for PLMs.}
In the fine-tuning stage, traditional fine-tuning paradigms introduce 
new parameters for task-specific predictions,
which could lead to the over-fitting problem in low-resource scenarios~\cite{Brown2020Language,Schick2021Exploiting}. 
To address the issue, 
prompt-tuning has recently been proposed~\cite{Brown2020Language,Schick2021Exploiting,Gao2021Making,Xiao2021GPT}. For example,
GPT-3~\cite{Brown2020Language} proposes to enable in-context learning with handcraft prompts in zero-shot scenarios.
In addition,
\citet{Schick2021Exploiting} and \citet{Gu2021PPT} explore all the tasks in a unified cloze-style format,
which boosts the performance of PLMs in few-shot learning tasks.
There are also recent works that 
improve prompt-tuning 
based on heuristic rules~\cite{han2021ptr}, automatic prompt construction~\cite{Gao2021Making,Shin2020AutoPrompt} and continuous prompt learning~\cite{Xiao2021GPT,Liu21PTuningv2,Hu2021Knowledgeable,Gu2021PPT}.
Inspired by prompt-tuning, 
we inject factual knowledge into PLMs by designing multiple continuous prompts.

\section{The~{\model} Framework}

In this section, we first describe the basic notations, and then formally present the techniques of the~{\model} in detail. 

\subsection{Notations}
A knowledge graph is denoted as $\mathcal{G}=(\mathcal{E}, \mathcal{R}, \mathcal{T})$.
Here, $\mathcal{E}$ is a set of entities,
$\mathcal{R}$ is a set of relations,
and $\mathcal{T}$ is a set of triples to express {semantic knowledge}. 
Specifically,
$\mathcal{T} = \{(e_h, r, e_t)| e_h, e_t\in\mathcal{E}, r\in\mathcal{R}\}$, where $e_h$, $r$ and $e_t$ denote head entity, relation and tail entity, respectively. 
Given a knowledge graph $\mathcal{G}$,
let $(e_0, \{(r_i, e_i)\}_{i=1}^k)$
be a $k$-hop relation path starting from entity $e_0$ to $e_k$ in $\mathcal{G}$.
We further denote 
$\mathcal{V}$ as the vocabulary set used in PLMs
and $S=\{w_1, w_2, \cdots\}$ as a sentence, 
where $w_i\in\mathcal{V}$ is the $i$-th token of the sentence. 




\begin{figure}
\centering
\includegraphics[width=\linewidth]{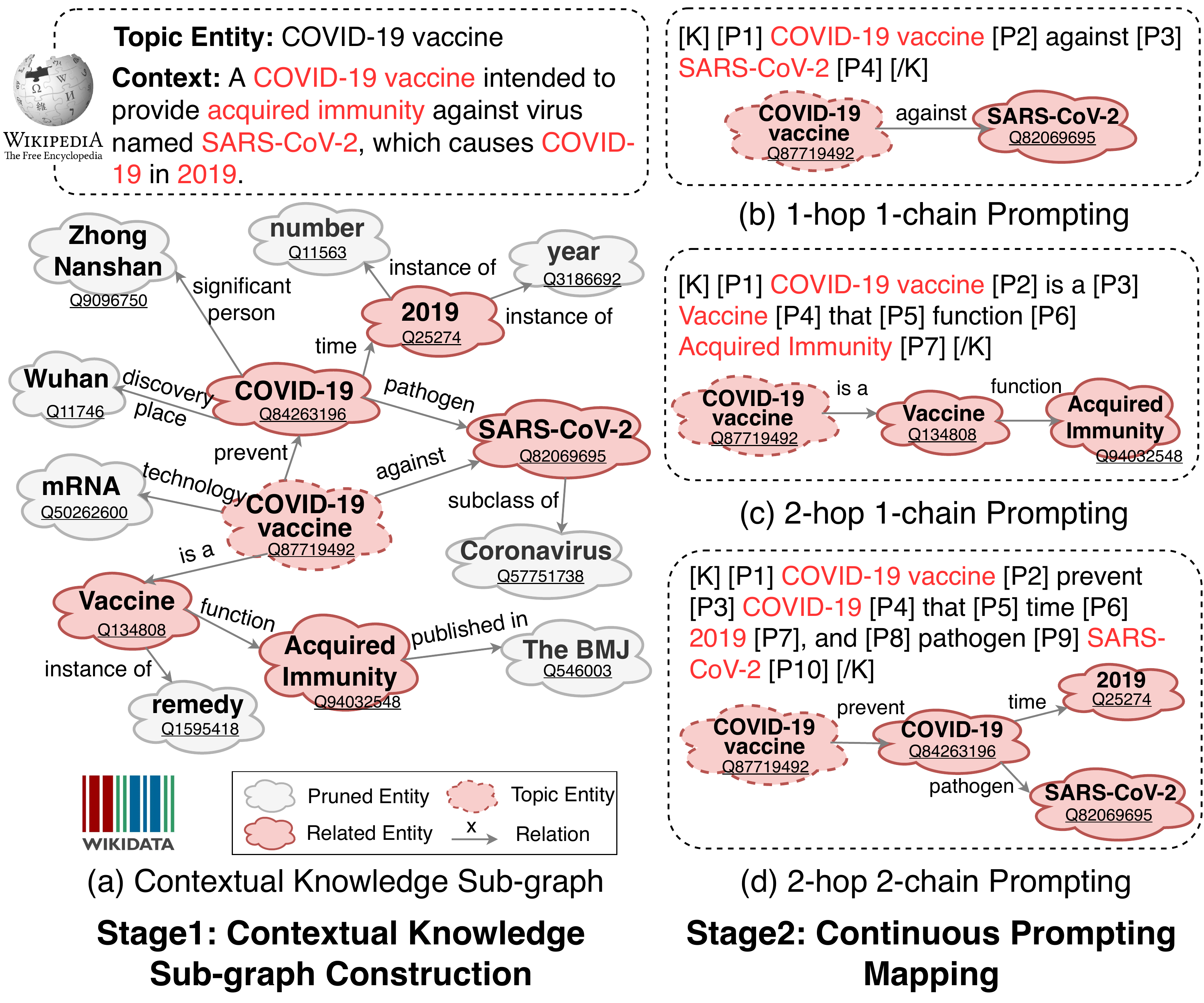}
\caption{The illustration on knowledge prompting. We first construct a 2-hop sub-graph for each context based on target entity, and then generate continuous prompts based on three kinds of mapping formats. (Best viewed in color.)}
\label{fig:prompt}
\end{figure}

\subsection{Knowledge Prompting}
\label{knowledge-prompting}
Knowledge prompting aims to
construct the knowledge sub-graph for each sentence, and then
transform factual knowledge into natural language prompts. 
Figure~\ref{fig:prompt}
illustrates the process in two steps.

\noindent\textbf{Contextual Knowledge Sub-graph Construction.}
Most existing methods integrate knowledge from KBs indiscriminately with all the mentions in the original context, 
which could bring in redundant and irrelevant information to PLMs~\cite{Zhang2021DKPLM}. 
Generally, 
a pre-training context is derived from the encyclopedia (e.g. Wikipedia), 
it is usually tagged with an entry that describes the topic of the context and can be viewed as the \emph{topic entity}.
Intuitively,
the sub-graph centering at the topic entity in KBs could contain semantic
knowledge that is highly correlated with the context.
Therefore, we propose to construct a KB sub-graph for each context based on the corresponding topic entity. 




We first scan all the entries in
Wikipedia Dumps
and derive their corresponding texts, which form a large-scale corpus.
After that,
for each sentence $S$, 
we utilize the TAGME
toolkit~\cite{Ferragina2010TAGME} to generate
$M_S$, a set of entity mentions in the sentence. 
Then,
for each sentence $S$ and 
its associated topic entity $e_S$,
we generate a 2-hop sub-graph 
$\mathcal{G}_S$ for $S$,
which centers at $e_S$ and includes all the $k$-hop relation paths starting from $e_S$ in $\mathcal{G}$. 
Here, $k\leq2$. 
However,
this will lead to identical sub-graphs for various sentences with the same topic entity.
To distinguish these sentences and filter irrelevant knowledge from the sub-graph,
we propose to further prune $\mathcal{G}_S$.
The procedure can be summarized as follows.
We first traverse all the 2-hop relation paths in $\mathcal{G}_S$. 
For each 2-hop relation path $(e_S, r_1, e_1, r_2, e_2)$,
if the tail entity $e_2$ is not mentioned in the sentence,
we remove $e_2$ and all the relations linked to it from $\mathcal{G}_S$.
After that,
we further traverse all the 1-hop relation paths in the pruned sub-graph.
For each 1-hop relation path $(e_S, r_1, e_1)$,
if $e_1$ is not mentioned in the sentence and there are not any 2-hop relation paths that contain $e_1$,
we remove $e_1$ and all the linked relations from the sub-graph.
For example, as shown in Figure~\ref{fig:prompt} (a), 
while the entity ``Vaccine'' is not a mention in the sentence,
we retain it 
due to its inclusion in the 2-hop relation path \emph{``(COVID-19 Vaccine, (is a, Vaccine), (function, Acquired Immunity))''}. 
Finally, we take the generated pruned sub-graph
as
the contextual knowledge sub-graph $\hat{\mathcal{G}_S}$ for $S$.

\begin{figure*}
\centering
\includegraphics[width=0.95\textwidth]{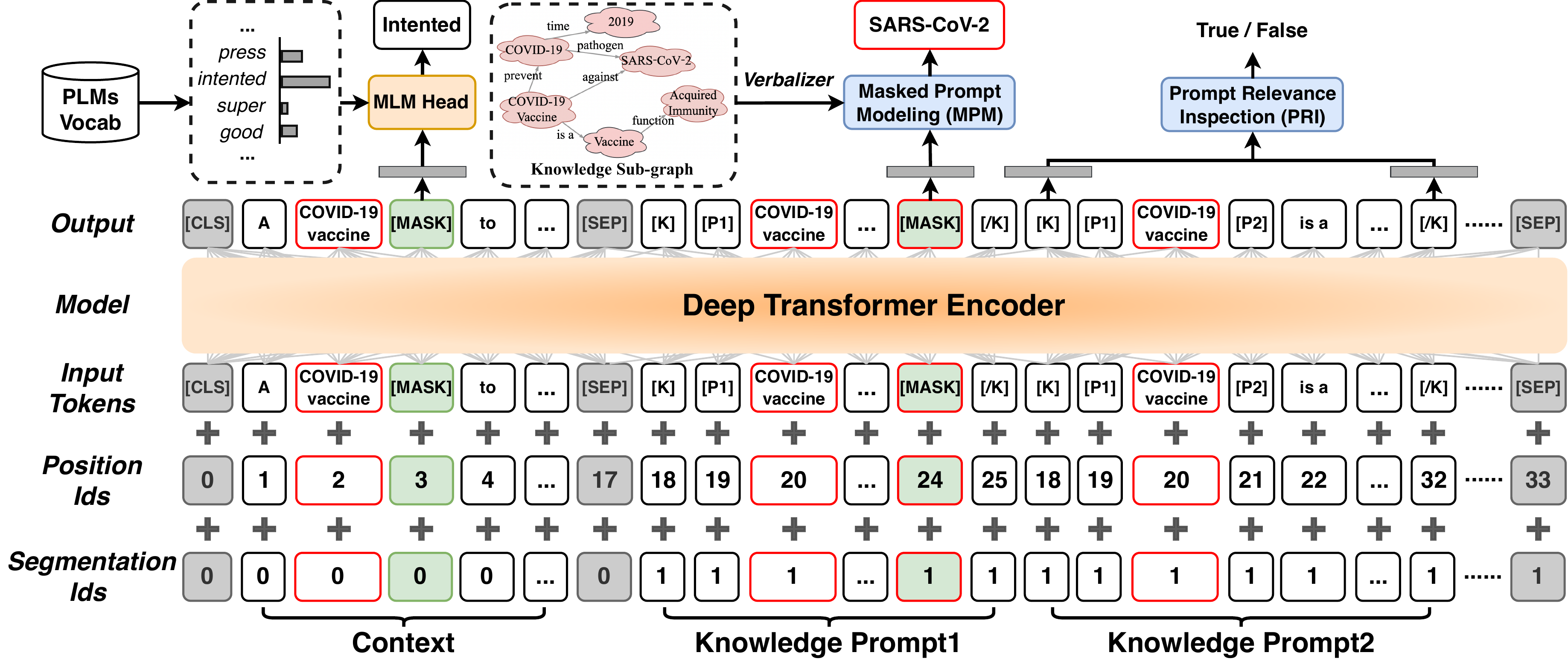}
\caption{The model architecture of~{\model} framework. (Best viewed in color.)
}
\label{fig:model}
\end{figure*}

\noindent\textbf{Continuous prompting mapping.}
After $\hat{\mathcal{G}_S}$ is constructed,
we next extract semantic knowledge from it. 
While there are some methods~\cite{Yao2019kgbert,Sun2020CoLAKE} that directly convert relation triples into discrete texts, 
it has been pointed out in~\cite{Xiao2021GPT} that the optimization in these methods could suffer from local minima.
Inspired by~\cite{Xiao2021GPT},
we employ continuous prompts
to inject factual knowledge into PLMs,
which have been shown to be effective in capturing semantic knowledge~\cite{Gu2021PPT}.

Specifically,
we design three types of prompt mapping rules based on the first-order and second-order structural information in $\hat{\mathcal{G}_S}$. 
As shown in Figure~\ref{fig:prompt},
the {1-hop 1-chain 
sub-structure}
is a single triple, 
the {2-hop 1-chain sub-structure} denotes a 2-hop relation path and 
the {2-hop 2-chain sub-structure} represents two 2-hop relation paths that share the same prefix triple.
These three types of sub-structures are the building blocks of other complex and 
higher-order ones, 
so they are adopted to design prompts.
Our framework can also be easily incorporated with other sub-structures.
For each type of sub-structure,
we design continuous templates with knowledge trigger tokens \texttt{[K]}, \texttt{[/K]} and pseudo tokens \texttt{[P$i$]}.
Examples of prompt templates are given in Figure~\ref{fig:prompt} (b)-(d).



Finally, we concatenate all the derived prompts with the original context to form an input sequence 
$X=\texttt{Concat}(S,\{P_i\}_{i=1}^m)$,
where $P_i$ denotes the $i$-th prompt and $m$ is the number of the prompts.



\subsection{Model Architecture}
After the input sequence $X$ is generated,
we first utilize the PLM tokenizer to transform $X$ into a token sequence $ \{x_j\}_{j=1}^n$, where $x_j$ is the $j$-th token and $n$ is the sequence length.
For notation simplicity,
we overload $X = \{x_j\}_{j=1}^n$.
Then
we feed $X$ into a PLM, 
whose architecture is shown in Figure~\ref{fig:model}.
We next describe the two major components.


\noindent\textbf{Input Embedding Layer.} 
Similar as previous methods~\cite{Liu2020KBERT, Sun2020CoLAKE},
there are three types of embeddings including token embeddings, position embeddings and segmentation embeddings.
For token embeddings, 
the trigger and pseudo tokens are randomly initialized 
while others 
are initialized by looking up the PLM embedding table. 
To alleviate the influence 
of the prompt order 
on the model performance,
we use the same segmentation id for all the prompts and
set the same position id to their start tokens.
For example,
in Figure~\ref{fig:model},
the segmentation ids of
all the prompts are set to 1 and 
the position ids of their start token
\texttt{[K]} are uniformly set to 18. 


\noindent\textbf{Deep Transformer Encoder.}
Following~\cite{Devlin2019BERT,Liu2019RoBERTa}, we use a multi-layer transformer encoder~\cite{Vaswani2017Attention} to capture the semantics from both context and knowledge prompts. 
To further mitigate the model's sensitivity 
to the prompt order, 
we design a mask matrix $\mathbf{M}$ for self-attention in PLMs,
whose $(i,j)$-th entry characterizes the relation between two tokens $x_i, x_j\in X$,
and is formally defined as:
\begin{equation}
\begin{aligned}
\mathbf{M}_{ij} = \left\{
        \begin{array}{rcl}
        -\inf  & & x_i\in P_u\land x_j\in P_v\land u\neq v; \\
        0     & & \text{otherwise}.\\
        \end{array} 
    \right.
\label{eqn:attention-mask}
\end{aligned}
\end{equation}
The self-attention in the transformer can then be calculated as:
\begin{equation}
\begin{aligned}
\text{Att}(\mathbf{Q}, \mathbf{K}, \mathbf{V}) = \texttt{Softmax}\big(\frac{\mathbf{Q}\mathbf{K}^\mathsf{T}}{\sqrt{d}} + \mathbf{M}\big)\mathbf{V},
\label{eqn:attention}
\end{aligned}
\end{equation}
where $d>0$ is the scale factor and $\mathbf{Q}, \mathbf{K}, \mathbf{V}\in\mathbb{R}^{n\times h}$ are the query, key and value matrices, respectively. 
Here,
$\mathbf{M}_{ij}=-\inf$ 
enforces that the attention weight between $x_i$ and $x_j$ from different prompts
will be set to 0.



\subsection{Self-supervised Pre-training Tasks}
We next provide the detailed description on two self-supervised pre-training tasks.



\noindent\textbf{Prompt Relevance Inspection (PRI).}
Since prompts can inject factual knowledge
into PLMs,
they are expected to be semantically
relevant to the context sequences.
Therefore,
we design a novel 
prompt relevance inspection
task,
which enhances the model's capability in learning the relevance of 
a prompt to a sentence.
For each sentence $S$ in the training corpus,
the knowledge prompting process  (see Section~\ref{knowledge-prompting}) can generate a set of relevant prompts 
$\mathcal{P}_S = \{P_i\}_{i=1}^m$ to $S$.
We then construct a ``positive'' prompt set $\texttt{Pos}$
by 
randomly selecting a prompt from $\mathcal{P}_S$ for each sentence $S$ in the corpus.
Further,
we need to construct a ``negative'' prompt set $\texttt{Neg}$.
For each sentence $S$,
we randomly select a prompt $P$ from $\mathcal{P}_S$ and replace a randomly selected entity in $P$
by an arbitrary entity in KBs.
The updated prompt is then labeled as a negative prompt and added to $\texttt{Neg}$.
We repeat the above process if 
more negative samples are needed.

After $\texttt{Pos}$ and $\texttt{Neg}$
are generated,
we can construct a training set $\mathcal{D}_{1}=\{(P,y_P)\}$,
where $y_P = 1$ if $P \in \texttt{Pos}$; 
$0$, otherwise.
For each sample $(P,y_P)$ from $\mathcal{D}_{1}$,
inspired by SpanBERT~\cite{Joshi2020SpanBERT},
we represent $P$ by using its two boundary tokens (\texttt{[K]} and \texttt{[/K]}).
Let $\mathbf{x}_{\texttt{[K]}},\mathbf{x}_{\texttt{[/K]}}\in\mathbb{R}^{h}$ be input embeddings, respectively.
Formally, we have:
\vspace{-1em}
\begin{equation}
\mathbf{h}_{P} = \mathcal{LN}(\sigma((\mathcal{F}_{\theta}(\mathbf{x}_{\texttt{[K]}}) + \mathcal{F}_{\theta}(\mathbf{x}_{\texttt{[/K]}}))\mathbf{W}_{1})),
\label{eqn:prompt-representation}
\end{equation}
where $\mathbf{W}_{1}\in\mathbb{R}^{h\times h}$ is the trainable parameter matrix, $\sigma(\cdot)$ and $\mathcal{LN}(\cdot)$ are the 
Sigmoid and LayerNorm functions, respectively. 
Further,
$\mathcal{F}_{\theta}(\cdot)$ is the output representation by the PLM and 
$\theta$ denotes the model parameters. 
Based on $\mathbf{h}_{P}$,
we define a binary classifier whose objective function is:
\begin{equation}
\mathcal{L}_{\text{PRI}} = \mathbb{E}_{(P, y_P)\sim\mathcal{D}_{1}}[ \log\text{Pr}_{\Phi}(y_P|P)],
\label{eqn:knowledge-digestion-loss}
\end{equation}
\vspace{-1em}
\begin{equation}
\begin{aligned}
\text{Pr}_{\Phi}(y_P|P) = \texttt{Softmax}(\mathcal{H}_{1}(\mathbf{h}_{P})),
\label{eqn:knowledge-digestion-pr}
\end{aligned}
\end{equation}
where $\mathcal{H}_{1}$ is the classification head with parameter $\Phi$ and $\text{Pr}(\cdot)$ denotes the probability distribution. 

\noindent\textbf{Masked Prompt Modeling (MPM).}
We further propose a masked prompt modeling task,
which aims to predict 
the masked entity in a prompt. 
Different from the 
vanilla masked language modeling~\cite{Devlin2019BERT}
that has an enormous search space over the whole PLM vocabulary $\mathcal{V}$,
our task shrinks the search space 
for a prompt $P \in \mathcal{P}_S$ 
to the entity set of
the contextual knowledge sub-graph $\hat{\mathcal{G}}_S$.

Given a training corpus,
for each context $S$,
we randomly mask an entity $e_P$ (except the topic entity\footnote{The topic entity is fixed at the start position in all the relation paths. We do not mask it because it is easy for PLMs to predict based on other knowledge prompts.}) with the \texttt{[MASK]} token in an arbitrarily selected prompt $P\in\mathcal{P}_S$.
Then we can generate a dataset
$\mathcal{D}_2=\{(P,e_P)\}$ that consists of all the (prompt, masked entity) pairs.
To enforce the model to better learn 
factual knowledge expressed by the generated prompts,
we employ contrastive learning~\cite{Jean2015On} techniques 
and formulate an objective function as:
\begin{equation}
\begin{aligned}
& \mathcal{L}_{\text{MPM}}=\mathbb{E}_{(P,e_P)\sim\mathcal{D}_2}\\
& \left [\frac{\exp(g(e_P|P))}{\exp(g(e_P|P)) + N\mathbb{E}_{e'_P\sim q(e)}\left [\exp(g(e'_P|P))\right]}\right],
\label{eqn:knowledge-prediction-loss}
\end{aligned}
\end{equation}
where 
$e'_P \neq e_P$ is a negative entity,
$N$ is the total number of sampled negative entities\footnote{If $|\mathcal{E}_S|<N+1$, the remaining $N-|\mathcal{E}_S|+1$ negative entities can be sampled from the whole $\mathcal{E}$.}, 
and
$q(\cdot)$ is a sampling function implemented by the PageRank algorithm~\cite{Gao2021SimCSE}, 
which is used to calculate the sampling probability scores for all the entities in $\mathcal{E}_S$. 
Further,
$g(\cdot)$ 
is a scoring function that
measures the similarity between the masked token representation $\mathbf{h}_{\texttt{[MASK]}}$ and the entity $e_P$.
In particular, 
$g(\cdot)$ is expected to effectively 
learn the representation of the masked entity $e_P$
by pulling embeddings of all its mentions in contexts
together 
and pushing apart that of other negative entities.
Therefore,
we first
introduce a verbalizer mapping function $\hat{f}(e_P)$,
which maps $e_P$ 
to the union of all its mention lists in the contexts of the corpus.
Then
following~\cite{Page1999The},
we compute:
\begin{equation}
\begin{aligned}
g(e_P|P) = \mathbb{E}_{v\sim \hat{f}(e_P)}\left [(\mathbf{h}_{\texttt{[MASK]}})^\mathsf{T}\mathbf{x}_{v}\right] - \log q(e_P)
\label{eqn:knowledge-prediction-loss2}
\end{aligned}
\end{equation}
\vspace{-1em}
\begin{equation}
\begin{aligned}
\mathbf{h}_{\texttt{[MASK]}} = \mathcal{LN}(\sigma(\mathcal{F}_{\theta}(\mathbf{x}_{\texttt{[MASK]}})\mathbf{W}_{2})).
\label{eqn:prompt-mask}
\end{aligned}
\end{equation}
Here,
$\mathbf{x}_{\texttt{[MASK]}}\in\mathbb{R}^{h}$ is the input embedding vector of the masked token and
$\mathbf{W}_{2}\in\mathbb{R}^{h\times h}$ is a trainable weight matrix. Further, 
$\mathbf{x}_{v}$ denotes the embedding of mention $v$ from $\hat{f}(e_P)$, 
which is calculated by 
averaging embeddings of 
all tokens included in $v$.





Finally, the total loss can be computed as:
\begin{equation}
\begin{aligned}
\mathcal{L} = \mathcal{L}_{\text{MLM}} + \lambda\mathcal{L}_{\text{PRI}} + \mu\mathcal{L}_{\text{MPM}},
\label{eqn:total-loss}
\end{aligned}
\end{equation}
where $\mathcal{L}_{\text{MLM}}$ denotes the vanilla MLM objective in PLMs, $\lambda, \mu\in[0,1]$ are the balancing coefficients.

\section{Experiments}

In this section, we comprehensively evaluate the effectiveness of {\model}.
We also conduct the hyper-parameter analysis in Appendix~\ref{app:ha}.

\subsection{Implementation Details}

Following previous works~\cite{Sun2020CoLAKE,Zhang2021DKPLM}, 
the pre-training data is collected from 
Wikipedia Dumps (2020/03/01)\footnote{\url{https://dumps.wikimedia.org/enwiki/}.} and consists of 25,933,196 sentences. Further, the KB used is WikiData5M~\cite{Wang2021KEPLER}, which includes $3,085,345$ entities and $822$ relation types. 

For the baselines, we select six knowledge-enhanced PLMs: 
1) \textbf{ERNIE-THU}~\cite{Zhang2019ERNIE} integrates knowledge embeddings with aligned mentions. 
2) \textbf{KnowBERT}~\cite{Peters2019Knowledge} 
utilizes the attention mechanism to realize knowledge fusion. 
3) \textbf{KEPLER}~\cite{Wang2021KEPLER} introduces a novel knowledge embedding loss for capturing knowledge. 4) \textbf{CoLAKE}~\cite{Sun2020CoLAKE} constructs 
a unified graph from context and knowledge base. 
5) \textbf{K-Adapter}~\cite{Wang2021KAdapter}
learns representations for different kinds of knowledge via neural adapters.
6) \textbf{DKPLM}~\cite{Zhang2021DKPLM} aims at injecting long-tailed entities.



\begin{table}
\centering
\begin{small}
\begin{tabular}{l | ccc}
\toprule
\bf Models &\bf  P &\bf  R &\bf  F1 \\
\midrule
UFET & 77.4 & 60.6 & 68.0 \\
BERT & 76.4 & 71.0 & 73.6 \\
RoBERTa & 77.4 & 73.6 & 75.4 \\
\midrule
ERNIE$_{BERT}$ & 78.4 & 72.9 & 75.6  \\
ERNIE$_{RoBERTa}$ & 80.3 & 70.2 & 74.9  \\
KnowBERT$_{BERT}$ & 77.9 & 71.2 & 74.4  \\
KnowBERT$_{RoBERTa}$ & 78.7 & 72.7 & 75.6  \\
KEPLER$_{WiKi}$ & 77.8 & 74.6 & 76.2  \\
CoLAKE & 77.0 & 75.7 & 76.4  \\
DKPLM & 79.2 & 75.9 & 77.5 \\ 
\midrule
{\model} & \textbf{80.8} & 75.1 & 77.8 \\ 
{\model}$_{KNOW}$ & 80.5 & \textbf{76.1} & \textbf{78.2} \\ 
\bottomrule
\end{tabular}
\end{small}
\caption{The results (\%) on Open Entity.}
\label{tab:entity-typing}
\end{table}


\begin{table}
\centering
\resizebox{\linewidth}{!}{
\begin{tabular}{l|ccc|ccc}
\toprule
\multirow{2}*{\bf Methods} &  \multicolumn{3}{c|}{\bf TACRED} & \multicolumn{3}{c}{\bf FewRel} \\
& \bf P & \bf R & \bf F1 & \bf P & \bf R & \bf F1 \\
\midrule
CNN & 70.3 & 54.2 & 61.2 & - & - & - \\
PA-LSTM & 65.7  & 64.5 & 65.1 & - & - & - \\
C-GCN & 69.90 & 63.3 & 66.4 & - & - & - \\
BERT & 67.2 & 64.8 & 66.0  & 84.9 & 85.1 & 85.0 \\
RoBERTa & 70.0 & 69.6 & 70.2 & 85.4 & 85.4 & 85.3 \\ \midrule
ERNIE$_{BERT}$ & 70.0 & 66.1 & 68.1  & 88.5 & 88.4 & 88.3 \\
KnowBERT & 71.6 & 71.5 & 71.5  & - & - & - \\
CoLAKE & 72.8 & 73.5 & 73.1 & \textbf{90.6}& \textbf{90.6} & \textbf{90.6} \\
DKPLM & 72.6 & 73.5 &  73.0  & - & - & - \\ 
\midrule
{\model} & 72.6 & 73.7 & 73.3  & 87.6 & 87.4 & 87.5 \\ 
{\model}$_{KNOW}$ & \textbf{73.3} & \textbf{73.9} & \textbf{73.5} & \textbf{88.9} & \textbf{88.9} & \textbf{88.8} \\ 
\bottomrule
\end{tabular}
}
\caption{The results (\%) on TACRED and FewRel.}
\label{tab:relation-extraction}
\end{table}

In the pre-training stage, 
we choose RoBERTa-base~\cite{Liu2019RoBERTa} from HuggingFace\footnote{\url{https://huggingface.co/transformers}.} as the default 
PLM.
Our framework can also be easily combined with other PLMs, such as BERT~\cite{Nina2019BERT} and DeBERTa~\cite{He2021Deberta}.
In the fine-tuning stage (if have), we fine-tune each task with only task-specific data. 
Further, 
we introduce a model variant~{\model}$_{KNOW}$, 
which 
employs knowledge prompts in the fine-tuning stage by
directly concatenating them with each example.
More implementation details about the pre-training and fine-tuning stages are provided in Appendices~\ref{appendix:pre-train-detail} and~\ref{appendix:task-specific-detail}, respectively.

\begin{table*}[t]
\centering
\resizebox{\linewidth}{!}{
\begin{tabular}{c|ccc|cccc|c}
\toprule
\bf Datasets & \bf  ELMo & \bf  BERT & \bf  RoBERTa &  \bf  CoLAKE & \bf  K-Adapter$^\ast$ & \bf  KEPLER & \bf  DKPLM &  \bf  {\model} 	 \\
\midrule
Google-RE & 2.2 & 11.4 & 5.3  & 9.5  & 7.0 & 7.3 & 10.8 & \textbf{11.0} \\
UHN-Google-RE & 2.3 & 5.7 & 2.2 & 4.9  & 3.7 & 4.1 & 5.4 & \textbf{5.6}  \\ \midrule
T-REx & 0.2  & 32.5 & 24.7 & 28.8  & 29.1 & 24.6 & 32.0 & \textbf{32.3}  \\
UHN-T-REx & 0.2 & 23.3 & 17.0 & 20.4 & \textbf{23.0}  & 17.1 & 22.9 & 22.5 \\ \bottomrule
\end{tabular}
}
\caption{The performance P@1 (\%) of knowledge probing. 
Besides, K-Adapter$^\ast$ is based on RoBERTa-large and uses a subset of T-REx as its training data, which may contribute to its superiority over other methods.}
\label{tab:knowledge-probing}
\end{table*}

\begin{table*}[ht]
\centering
\resizebox{\linewidth}{!}{
\begin{tabular}{c | c | llllllll | c}
\toprule
\multirow{3}*{\bf Paradigms} &
\multirow{3}*{\bf Methods} &
\multicolumn{8}{c|}{\emph{Single-Sentence Natural Language Understanding Tasks}} & \multirow{3}*{\bf Avg.} \\
& & \bf SST-2 & \bf SST-5 & \bf MR & \bf CR & \bf MPQA & \bf Subj & \bf TREC & \bf CoLA & \\
&  & (acc) & (acc) & (acc) & (acc) & (acc) & (acc) & (acc) & (matt.) & \\
\midrule
\multirow{2}*{PT-Zero} & RoBERTa & 82.57	& 29.46 & \textbf{65.10} & \textbf{82.15} & 49.90 & \textbf{69.20} & 20.80 & -4.89 & 49.29 \\
&{\model} & \textbf{84.15} &	\textbf{30.67} &	64.15 &	81.60 &	\textbf{53.80} &	68.70 &	\textbf{24.80} &	\textbf{-2.99} &	\textbf{50.61} \\
\cmidrule(r){1-11}
\multirow{2}*{PT-Few} & RoBERTa & 86.35\small{\textpm1.3} &	36.79\small{\textpm2.0} &	\textbf{83.35}\small{\textpm0.9} &	\textbf{88.85}\small{\textpm1.4} &	66.40\small{\textpm1.9} &	89.25\small{\textpm2.6} &	76.80\small{\textpm5.0} &	6.61\small{\textpm6.9} & 66.80 \\
& {\model} &	\textbf{90.71}\small{\textpm1.0} &	\textbf{44.21}\small{\textpm2.9} &	82.00\small{\textpm1.5} &	85.35\small{\textpm0.4} &	\textbf{67.30}\small{\textpm1.2} &	\textbf{91.45}\small{\textpm0.4} &	\textbf{81.00}\small{\textpm3.3} &	\textbf{24.28}\small{\textpm11.3} &	\textbf{70.79} \\
\cmidrule(r){1-11}
\multirow{2}*{FT-Full} & RoBERTa &	94.90 &	56.90 &	\textbf{89.60} &	88.80 &	86.30 &	\textbf{96.50} &	\textbf{97.10} &	63.90 &	84.25 \\
& {\model} &	\textbf{95.30} &	\textbf{57.63} &	89.20 &	\textbf{89.10} &	\textbf{87.40} &	96.20 &	\textbf{97.10} &	\textbf{64.87} &	\textbf{84.60} \\

\bottomrule
\toprule

\multirow{3}*{\bf Paradigms} &
\multirow{3}*{\bf Methods} &
\multicolumn{8}{c|}{\emph{Sentence-Pair Natural Language Understanding Tasks}} & \multirow{3}*{\bf Avg.} \\
& & \bf MNLI & \bf MNLI-mm & \bf SNLI & \bf QNLI & \bf RTE & \bf MRPC & \bf QQP & \bf STS-B & \\
&  & (acc) & (acc) & (acc) & (acc) & (acc) & (f1) & (f1) & (pear.) & \\
\midrule
\multirow{2}*{PT-Zero} & RoBERTa & 38.92 &	39.09 &	\textbf{36.33} &	\textbf{47.08} &	55.96 &	54.87 &	45.72 &	-4.10 &	39.23 \\
& {\model} & \textbf{40.10} &	\textbf{43.53} &	36.23 &	46.87 &	\textbf{58.29} &	\textbf{58.97} &	\textbf{46.30} &	\textbf{-1.30} &	\textbf{41.12} \\
\cmidrule(r){1-11}
\multirow{2}*{PT-Few} & RoBERTa & 50.87\small{\textpm1.8} &	53.01\small{\textpm2.2} &	64.96\small{\textpm1.9} &	\textbf{60.08}\small{\textpm2.4} &	63.54\small{\textpm4.0} &	78.57\small{\textpm3.7} &	45.72\small{\textpm3.4} &	52.26\small{\textpm7.0} &	58.63 \\
& {\model} &	\textbf{55.50}\small{\textpm1.1} &	\textbf{57.56}\small{\textpm2.1} &	\textbf{66.33}\small{\textpm1.9} &	58.75\small{\textpm2.2} &	\textbf{66.25}\small{\textpm3.8} &	\textbf{79.78}\small{\textpm3.1} &	\textbf{62.60}\small{\textpm4.0} &	\textbf{68.79}\small{\textpm4.2} &	\textbf{64.45} \\
\cmidrule(r){1-11}
\multirow{2}*{FT-Full} & RoBERTa &	87.33 &	87.01 &	\textbf{92.10} &	92.58 &	77.32 &	90.23 &	92.16 &	91.12 &	88.73 \\
& {\model} &	\textbf{87.99} &	\textbf{87.32} &	92.01 &	\textbf{93.04} &	\textbf{79.48} &	\textbf{91.81} &	\textbf{92.42} &	\textbf{91.60} &	\textbf{89.46} \\

\bottomrule
\end{tabular}
}
\caption{The comparison between~{\model} and RoBERTa-base~\cite{Liu2019RoBERTa} over multiple natural language understanding (NLU) tasks in terms of acc/f1/matt./pear. (\%) and standard deviation with three paradigms, such as zero-shot prompt-tuning (PT-Zero), few-shot prompt-tuning (PT-Few) and full-data fine-tuning (FT-Full).}
\label{tab:nlu}
\end{table*}

\subsection{Knowledge-aware Tasks}
In the fine-tuning stage, we use three tasks: entity typing, relation extraction and knowledge probing, to evaluate the model performance.

\noindent\textbf{Entity Typing.}
Given a sentence and a corresponding entity mention, the task is to 
predict the type of the 
mention. 
We choose precision (P), recall (R) and F1 score as the evaluation metrics.
For all these metrics, the larger the value, the better the model performance.
For fairness, 
we follow the same training settings as in~\cite{Zhang2021DKPLM} and 
use the Open Entity~\cite{Choi2018Ultra} dataset.
To show the effectiveness and competitiveness of our pre-trained model,
we also choose UFET~\cite{Choi2018Ultra}, BERT~\cite{Nina2019BERT} and RoBERTa~\cite{Liu2019RoBERTa} as traditional baselines for this task.

The results are summarized in Table~\ref{tab:entity-typing}.
From the table, 
we see that knowledge-enhanced PLMs generally perform better than traditional methods.
In particular,
our methods can achieve the best results w.r.t. all the metrics.
For example,   
the F1 score of~{\model}$_{KNOW}$ is 78.2\%, 
which improves that of the runner-up by 0.7\%. 
Further,
we also find that 
both~{\model} and~{\model}$_{KNOW}$ outperform RoBERTa.
This indicates that 
injecting knowledge prompts into both pre-training and fine-tuning stages are useful for improving the model performance on this task. 

\noindent\textbf{Relation Extraction.}
It aims to classify the relation between two given entities based on the corresponding texts. 
We follow~\cite{Sun2020CoLAKE} to choose two widely used tasks: TACRED~\cite{Zhang2017Position} and FewRel~\cite{han2018fewrel}.
Similar as in entity typing,
we take precision (P), recall (R) and F1 as the evaluation metrics.
We also compare our models with five traditional models including CNN, PA-LSTM~\cite{Zhang2017Position}, C-GCN~\cite{Zhang2018Graph}, BERT~\cite{Nina2019BERT} and RoBERTa~\cite{Liu2019RoBERTa}.

\begin{table*}[ht]
\centering
\begin{small}
\begin{tabular}{cc | ccc | ccc}
\toprule
\multirow{3}*{\bf Knowledge Paradigms} &
\multirow{3}*{\bf PLM} &
\multicolumn{6}{c}{\emph{Full-data Fine-tuning}} 
\\

& & \bf Open Entity & \bf TACRED & \bf FewRel & \bf SST-2 & \bf CoLA & \bf QQP \\
&  & (f1) & (f1) & (f1) & (acc) & (matt.) & (f1) \\
\midrule
\multirow{3}*{None} & BERT & 73.6 & 66.0 & 85.0 & 93.5 & 52.1 & 71.2 \\
& RoBERTa & 75.4 & 70.2 & 85.3 & 94.9 & 63.9 & 92.2 \\
& DeBERTa & 76.5 & 72.1 & 87.0 & 95.1 & 64.9 & 89.3 \\
\cmidrule(r){1-8}
\multirow{3}*{Knowledge Masking} & BERT & 74.4 & 70.5 & 85.8 & 93.9 & 51.6 & 71.4 \\
& RoBERTa & 75.8 & 71.3 & 86.0 & 95.0 & 64.3 & 92.0 \\
& DeBERTa & 77.0 & 72.6 & 86.8 & 94.7 & 65.0 & 90.3 \\
\cmidrule(r){1-8}
\multirow{3}*{Knowledge Fusion$^{\dag}$} & BERT & 75.6 & 68.1 & 88.3 & 93.5 & 52.3 & 71.2  \\
& RoBERTa & 74.9 & 68.4 & 88.4 & 93.9 & 63.6 & 91.5 \\
& DeBERTa & 76.8 & 70.6 & 88.8 & 94.2 & 65.7 & 89.9  \\
\cmidrule(r){1-8}
\multirow{3}*{Graph Learning$^{\ddag}$} & BERT & 75.1 & \textbf{72.9} & \textbf{88.4} & 94.2 & 53.9 & 72.0 \\
& RoBERTa & 76.4 & 73.1 & \textbf{90.6} & 94.6 & 63.4 & \textbf{93.3} \\
& DeBERTa & 77.1 & 72.8 & \textbf{89.5} & 94.0 & 66.1 & \textbf{92.2} \\
\cmidrule(r){1-8}
\multirow{3}*{{Knowledge Prompting}} & BERT & \textbf{76.0} & 72.2 & 87.1 & \textbf{94.6} & \textbf{57.3} & \textbf{75.8} \\
& RoBERTa & \textbf{77.8} & \textbf{73.3} & 87.5 & \textbf{95.3} & \textbf{64.9} & 92.4 \\
& DeBERTa & \textbf{77.7} & \textbf{73.5} & 88.0 & \textbf{95.6} & \textbf{66.3} & \textbf{92.2} \\
\bottomrule
\end{tabular}
\end{small}
\caption{The comparison between~{\model} and other knowledge-enhanced paradigms on different base PLMs. 
For each base PLM, 
we highlight the largest score in bold.
}
\label{tab:paradigm-comparison}
\end{table*}

As shown in Table~\ref{tab:relation-extraction}, 
most knowledge-enhanced PLMs outperform traditional methods by a large margin. 
This shows the necessity of external knowledge incorporation for PLMs.
For the TACRED task,
our models 
perform the best.
For the FewRel task,
while 
CoLAKE obtains the best results,
{\model}$_{KNOW}$
can improve the F1 score over ERNIE and RoBERTa by 0.5\% and 3.5\%, respectively.
In addition, the outperformance of {\model}$_{KNOW}$ over \model\ shows the importance of injecting knowledge prompts in the fine-tuning stage. 

\noindent\textbf{Knowledge Probing.}
Knowledge probing aims to evaluate whether the PLM possesses the intrinsic factual knowledge in zero-shot settings.
We compare all the methods w.r.t. P@1.
We select LAMA~\cite{Petroni2019Language} and the advanced version LAMA-UHN~\cite{Nina2019BERT} tasks with four datasets, including Google-RE, UHN-Google-RE, T-REx, and UHN-T-REx. 

From Table~\ref{tab:knowledge-probing},
our model~{\model} beats other knowledge-enhanced PLMs on Google-RE, UHN-Google-RE and T-REx datasets. 
For UHN-T-REx,
\model\ 
performs comparably with the winner 
and 
improve RoBERTa by 5.5\% in terms of P@1. 
In addition,
we see that BERT achieves the best performance over multiple tasks.
This is because it uses a small vocabulary set, 
which has also been pointed out in~\cite{Sun2020CoLAKE}.

\subsection{Performance on General NLU Tasks}

We further investigate whether {knowledge prompting} can consistently improve the PLM performance in 
both full-resource and low-resource learning.
We follow~\cite{Gao2021Making} to select 15 widely used NLU tasks.
We consider three training settings, 
including zero-shot prompt-tuning (PT-Zero), few-shot prompt-tuning (PT-Few), and full data fine-tuning (FT-Full).
Details on these datasets and training procedures are provided in Appendix~\ref{appendix:nlu-task}. 

From Table~\ref{tab:nlu}, 
we make the following observations: 
1)~{\model} achieves the best overall results over all the training settings. 
The much larger margins of \model\ over others in 
PT-Zero and PT-Few 
show that continual pre-training by {knowledge prompting} indeed improves the performance when using the prompt-based fine-tuning technique. 
2) In the low-resource scenarios, the performances of both RoBERTa and~{\model} on single-sentence tasks are better than sentence-pair tasks, 
which indicates that the sentence-pair task is more difficult in the few-shot settings. 
3) We also find~{\model} sometimes performs worse than RoBERTa on MR and CR datasets. 
We conjecture 
that this is because these tasks are sensitive to external knowledge.

\begin{table}
\centering
\resizebox{\linewidth}{!}{
\begin{small}
\begin{tabular}{l | ccc}
\toprule
\bf Models &\bf  Open Entity &\bf  TACRED &\bf  FewRel \\
\midrule
\textbf{\model} & \bf 77.8 & \bf 73.3 & \bf 87.5 \\
\midrule
w/o. PRI & 77.5 & 72.8 & 87.1 \\
w/o. MPM & 77.4 & 72.5 & 86.8 \\
w/o. PRI \& MPM & 77.2 & 72.4 & 86.6 \\
w/o. cont. & 77.7 & 73.1 & 87.3 \\
w/o. mask. & 77.5 & 72.9 & 87.1 \\
w/o. all & 76.3 & 70.8 & 85.9 \\
\bottomrule
\end{tabular}
\end{small}
}
\caption{The ablation study results (F1 score \%). 
w/o. PRI removes the PRI task,
w/o. MPM removes the MPM task, 
w/o. PRI \& MPM removes both the PRI and MPM tasks,
w/o. cont. removes all continuous pseudo tokens, w/o. mask. replaces the mask matrix in Eq.~\ref{eqn:attention-mask} with the default in RoBERTa, and w/o. all denotes to remove all the proposed techniques.}
\label{tab:ablation}
\end{table}

\subsection{Knowledge Prompting Study}

We end this section with a further comparison between 
knowledge prompting 
and
other knowledge-enhanced paradigms
including knowledge masking, knowledge fusion and graph learning. 
For each paradigm, 
we employ three PLMs:
BERT-base, RoBERTa-base and DeBERTa-base.
For knowledge masking, we 
mask all the entity mentions in each training sentence, 
and train the model by the vanilla MLM objective.
For 
knowledge fusion and graph learning, 
we directly 
run the source codes of ERNIE
and CoLAKE,
respectively.
The results are summarized in Table~\ref{tab:paradigm-comparison}.
From the table,
we see that,
compared with other knowledge-enhanced paradigms,
for each base PLM,
knowledge prompting
can lead to largest performance gains in most downstream tasks.
This further
verifies the effectiveness of knowledge prompting in 
boosting the performance of existing PLMs.

\subsection{Ablation Study}
We conduct an ablation study to investigate the characteristics of main components in~{\model},
including prompt relevance inspection (PRI), masked prompt modeling (MPM), continuous pseudo tokens in the prompt and the mask matrix for self-attention in PLMs (Eq.~\ref{eqn:attention-mask}).
Table~\ref{tab:ablation} reports the F1 scores on the Open Entity, TACRED, and FewRel tasks. 
From the table,
we observe that
the removal of each component leads to the performance drop of \model, 
which shows the importance of all these components in~{\model}.

\section{Conclusion}
In this paper, 
we presented 
a seminal knowledge prompting paradigm,
based on which 
a novel knowledge-prompting-based PLM framework \model\ was proposed.
We constructed contextual knowledge sub-graphs for contexts and employed continuous prompting mapping to generate knowledge prompts.
After that,
we designed
two self-supervised pre-training tasks to learn semantic knowledge from prompts.
Finally,
we conducted extensive experiments to 
evaluate the model performance.
Experimental results validate 
the effectiveness of knowledge prompting in boosting the performance of PLMs.



\section*{Limitations}
We have listed some limitations:
1) In the experiments, we follow most previous works to only focus on 
general natural language understanding (NLU) tasks. We do not evaluate the performance over some question answering tasks (e.g. SQuAD, HotpotQA, etc.), we let it as the future research.
2) Our work focuses on the PLM without any  transformer decoders. We think it is possible to extend our method in natural language generation (NLG) tasks.

\section*{Ethical Considerations}


Our contribution in this work is fully methodological, namely a knowledge-prompting-based pre-trained language model ({\model}) to boost the performance of PLMs with factual knowledge.
Hence, there is no explicit negative social influences in this work.
However, transformer-based models may have some negative impacts, such as gender and social bias. 
Our work would unavoidably suffer from these issues.
We suggest that users should carefully address potential risks 
when the {\model} models are deployed online.




\section*{Acknowledgement}

Xiang Li would like to acknowledge the support from Shanghai Pujiang Talent Program (Project No. 21PJ1402900). 
This work has also been supported by the National Natural Science Foundation of China under Grant No. U1911203, 
Alibaba Group through the Alibaba Innovation Research Program, 
and the National Natural Science Foundation of China under Grant No. 61877018,
The Research Project of Shanghai Science and Technology Commission (20dz2260300) and The Fundamental Research Funds for the Central Universities.


\bibliography{anthology,custom}
\bibliographystyle{acl_natbib}

\appendix

\section{Details of the Pre-training}
\label{appendix:pre-train-detail}

\subsection{Training Corpus and Knowledge Base}

We follow~\cite{Liu2020KBERT,Sun2020CoLAKE,Zhang2021DKPLM} to collect training corpora from Wikipedia (2020/03/01)\footnote{\url{https://dumps.wikimedia.org/enwiki/}.}, and use WikiExtractor\footnote{\url{https://github.com/attardi/wikiextractor}.} to process the training data.
The knowledge base (KB) $\mathcal{G}$ we choose is WikiData5M~\cite{Wang2021KEPLER}, which is an urge-large structure data source based on Wikipedia.
The entity linking toolkit and the verbalizer we used are both the TAGME\footnote{\url{https://sobigdata.d4science.org/group/tagme}.}~\cite{Ferragina2010TAGME}, which can be viewed as a well-designed mapping function between KB entities set $\mathcal{E}$ and the PLM vocabulary set $\mathcal{V}$.
In total, we have 3,085,345 entities and 822 relation types in $\mathcal{G}$, and 25,933,196 training sentences.
For each sentence, we construct the contextual knowledge sub-graph and generate continuous knowledge prompts offline by the process described in Section~\ref{knowledge-prompting}.
In average, the number of the entities and knowledge prompts for each sentence are 15 and 8, respectively.

As mentioned above, our framework consists of three main training objectives. For the Masked Language Modeling (MLM) task, we follow~\cite{Devlin2019BERT, Liu2019RoBERTa} to randomly select 15\% tokens only in the context. For the selected tokens, 80\% of them are replaced with \texttt{[MASK]} token, and 10\% of them are replaced with randomly sampled tokens from the whole vocabulary set. For the Prompt Relevance Inspection (PRI) task, we only select one prompt from each input example to make the decision. For the Masked Prompt Modeling (MPM) task, we also only select one prompt (not equal to the selected prompt in PRI) and mask one entity.

\subsection{Pre-training Implement Details}

In the pre-training stage, we choose RoBERTa-base~\cite{Liu2019RoBERTa} from the HuggingFace\footnote{\url{https://huggingface.co/transformers/index.html}.} as the default underlying PLM. 
We train our model by AdamW algorithm with $\beta_1=0.9, \beta_2=0.98$. The learning rate is set as 1e-5 with a warm up rate 0.1. 
The best balance coefficients we found are $\lambda=\mu=0.5$. The number of negative entities is $N=10$. Especially, if the number of the entities in the current sub-graph $|\mathcal{E}_S|<N+1$, the remaining $N-|\mathcal{E}_S|+1$ negative entities can be sampled from the whole entities set $\mathcal{E}$ from KB $\mathcal{G}$.
We also leverage dropout and regularization strategies to avoid over-fitting. 
The model is trained on 8 V100-32G GPUs for 2.5 days with a total batch size of 768. 


\begin{table}
\centering
\resizebox{\linewidth}{!}{
\begin{small}
\begin{tabular}{l | cccc}
\toprule
\bf Datasets &\bf  \#Train &\bf  \#Develop &\bf  \#Test & \bf \#Class \\
\midrule
Open Entity & 2,000 & 2,000 & 2,000 & 6 \\
TACRED & 68,124 & 22,631 & 15,509 & 42 \\
FewRel & 8,000 & 8,000 & 16,000 & 80 \\
\bottomrule
\end{tabular}
\end{small}
}
\caption{The statistics of Open Entity, TACRED and FewRel.}
\label{tab:knowledge-aware-datasets}
\end{table}

\section{Details of the Task-specific Fine-tuning}
\label{appendix:task-specific-detail}

We choose multiple natural language processing (NLP) tasks for the model evaluations, including knowledge-aware tasks and general natural language understanding (NLU) tasks. In this section, we provide the details of the dataset information and fine-tuning procedures for each task.

\subsection{Knowledge-aware Tasks}

We select three knowledge-aware tasks, such as entity typing, relation extraction, and knowledge probing. The dataset statistics can be found in Table~\ref{tab:knowledge-aware-datasets}.

\noindent\textbf{Entity typing.}
Entity typing requires to classify the designated entity mention based on the given sentence. We select the Open Entity dataset for evaluation. It consists of 6 entity types, and 2,000 sentences for training, developing, and testing data, respectively.

The variant method~{\model}$_{KNOW}$ means adding injected knowledge prompts for each data. In this manner, the topic entity is the designated entity mention, we follow the proposed \emph{knowledge prompting} procedure to obtain the knowledge prompts for each sentence, and directly concatenate them with the original sentence. In the fine-tuning stage, we do not use knowledge-aware pre-training objectives.

During fine-tuning, we follow~\cite{Sun2020CoLAKE} to add two special tokens \texttt{[ENT]} and \texttt{[/ENT]} before and after the entity mention. We directly concatenate their representations at the top of the PLM and use them for classification. The max sequence length we set is 128, the learning rate is set as 2e-5 with a warm up rate of 0.1, and the batch size is set to 16. Generally, we can obtain the best performance after 5 epochs.

\noindent\textbf{Relation Extraction.}
Relation extraction is one of the significant tasks for information retrieval. It aims to classify the relation class between two given entities based on the corresponding aligned sentence. We select two widely used datasets, i.e. TACRED~\cite{Zhang2017Position} and FewRel~\cite{han2018fewrel}. TACRED has 42 relation types and 106, 264 sentences in total. FewRel is constructed by distant supervised~\cite{mintz2009distant} and denoised by human annotations. It consists of 320,000 sentences and 80 relations.

\begin{table*}[t]
\begin{center}
\centering
\resizebox{\linewidth}{!}{%
\begin{tabular}{clcrrrcl}
\toprule
\bf Category & \bf Dataset & \bf \#Class & \bf \#Train & \bf \#Test & \bf Type & \bf Labels (classification tasks) \\
\bottomrule
 & SST-2 & 2 & 6,920 & 872 & sentiment & positive, negative \\
& SST-5 & 5 & 8,544 & 2,210 & sentiment & v. pos., positive, neutral, negative, v. neg. \\
& MR & 2 & 8,662& 2,000 & sentiment & positive, negative \\
single- & CR & 2 & 1,775 & 2,000 & sentiment & positive, negative \\
sentence & MPQA & 2 & 8,606 & 2,000 & opinion polarity & positive, negative \\
& Subj & 2 & 8,000 & 2,000 & subjectivity & subjective, objective \\
& TREC & 6 & 5,452 & 500 & question cls. & abbr., entity, description, human, loc., num.\\
& CoLA & 2 & 8,551 & 1,042 & acceptability & grammatical, not\_grammatical\\
\midrule
 & MNLI & 3 & 392,702 & 9,815 & NLI & entailment, neutral, contradiction\\
& SNLI & 3 &  549,367 & 9,842 & NLI & entailment, neutral, contradiction \\
sentence- & QNLI & 2  & 104,743 & 5,463 & NLI & entailment, not\_entailment \\
pair & RTE & 2 & 2,490 & 277 & NLI &  entailment, not\_entailment \\
 & MRPC & 2  & 3,668 & 408 & paraphrase & equivalent, not\_equivalent \\
& QQP & 2 & 363,846 & 40,431 & paraphrase & equivalent, not\_equivalent  \\
& STS-B & $\mathbb{R}$  & 5,749 & 1,500  & sent. similarity & - \\
\bottomrule
\end{tabular}
}
\end{center}
\caption{The statistics of multiple NLU datasets. STS-B is a real-valued regression task over the interval $[0, 5]$). Since the original test data is invisible, we use the develop set as our test set.}
\label{tab:nlu-datasets}
\end{table*}

For the variant~{\model}$_{KNOW}$, following the procedure of \emph{knowledge prompting}, we construct a knowledge sub-graph for each sentence. Since each sentence consists of two entities, both of them can be viewed as the topic entity. Hence, we obtain two sub-graphs for each sentence and then merge two sub-graphs by removing the duplicate relation paths. After that, we construct multiple prompts and concatenate them with the original sentence. 

During fine-tuning, we follow~\cite{Sun2020CoLAKE} to add four special tokens \texttt{[HD]}, \texttt{[/HD]}, \texttt{[TL]} and \texttt{[/TL]} to identity the head entity and tail entity in the given sentence. We obtain the representations of \texttt{[HD]} and \texttt{[TL]} from the last layer of the PLM, and concatenate them to perform classification. For TACRED~\cite{Zhang2017Position} dataset, the max sequence length is 256, the learning rate is 3e-5 with a warm up rate 0.1, the batch size we set is 32. We run experiment for 5 epochs. For FewRel~\cite{han2018fewrel}, the max sequence length is 256, the learning rate is 2e-5 with a warm up rate 0.1, the batch size we set is 32. We run experiment for 8 epochs. Notice, we do not use the original $N$-way $K$-shot settings for FewRel~\cite{han2018fewrel}.

\noindent\textbf{Knowledge Probing.}
Knowledge probing is a challenging task to evaluate whether the PLM grasps factual knowledge. The task requires no training stages which belongs to the zero-shot setting. We select LAMA~\cite{Peters2019Knowledge} and LAMA-UHN~\cite{Nina2019BERT} tasks with four datasets, i.e., Google-RE, UHN-Google-RE, T-REx, and UHN-T-REx. The dataset statistics can be found in the original paper in~\cite{Nina2019BERT}.

During the evaluation, we follow~\cite{Zhang2019ERNIE} to directly use the given designed template. Specifically, given a head entity, a relation, and a template with \texttt{[MASK]} token, we feed the input sequence into the PLM and obtain the predicted result at the masked position. For example in Google-RE, a manually defined template of relation ``place\_of\_birth'' is ``\texttt{[S]} was born in \texttt{[MASK]}.'', we can denote \texttt{[S]} as the head entity ``Obama'', and let the PLM generate the answer ``US.'' at the position of \texttt{[MASK]}.

\subsection{General Natural Language Understanding Tasks}
\label{appendix:nlu-task}

We select NLU tasks for the evaluation in both full-resource and low-resource scenarios. We follow~\cite{Gao2021Making} to select 15 tasks include 8 tasks from GLUE benchmark~\cite{Wang2019GLUE}, SNLI~\cite{Bowman2015a}, and 6 other sentence-level classification tasks. The statistics of each dataset are shown in Table~\ref{tab:nlu-datasets}.

As mentioned above, we provide three training settings, including zero-shot prompt-tuning (PT-Zero), few-shot prompt-tuning (PT-Few) and full-data fine-tuning (FT-Full). We summarize the detail settings in the following.

\noindent\textbf{Prompt-tuning.}
For the settings of PT-Zero and PT-Few, we follow~\citet{Gao2021Making} to evaluate our framework based on the prompt-tuning method. Specifically, for each task, we directly use the automatically generated discrete template and verbalizer, where the discrete template denotes the token sequence with a single \texttt{[MASK]} token, and the verbalizer is the label mapping function that can map generated result token to the label. The template and verbalizer for each task we used can be found in Table E.1 in the original paper of LM-BFF~\cite{Gao2021Making}. Take sentiment analysis as an example, for the sentence $S=$``The film is very attractive.'', we have:
\begin{equation*}
    \resizebox{.85\hsize}{!}{%
    $X = \texttt{[CLS]} S~\text{It was}~\texttt{[MASK]}. \texttt{[SEP]}$
    }
\end{equation*}
\begin{equation*}
    \resizebox{\hsize}{!}{%
    $\mathcal{M} = \{\text{``positive'': ``great''}; \text{``negative'': ``boring''}\}$
    }
\end{equation*}
where $X$ is the input sequence, $\mathcal{M}(y)$ denotes the verbalizer that map the label $y$ to a label word. For example, $\mathcal{M}(\text{``positive''})=\text{``great''}$.
Under these settings, we can transform arbitrary NLU tasks into a cloze-style problem.
In our experiments, the baseline is RoBERTa-base~\cite{Liu2019RoBERTa}. In addition, to validate the contributions of our proposal, we do not use other techniques mentioned in LM-BFF, such as demonstration learning and prompt ensemble.


\begin{figure}[t]
\centering
\begin{tabular}{ll}
\begin{minipage}[t]{0.48\linewidth}
    \includegraphics[width = 1\linewidth]{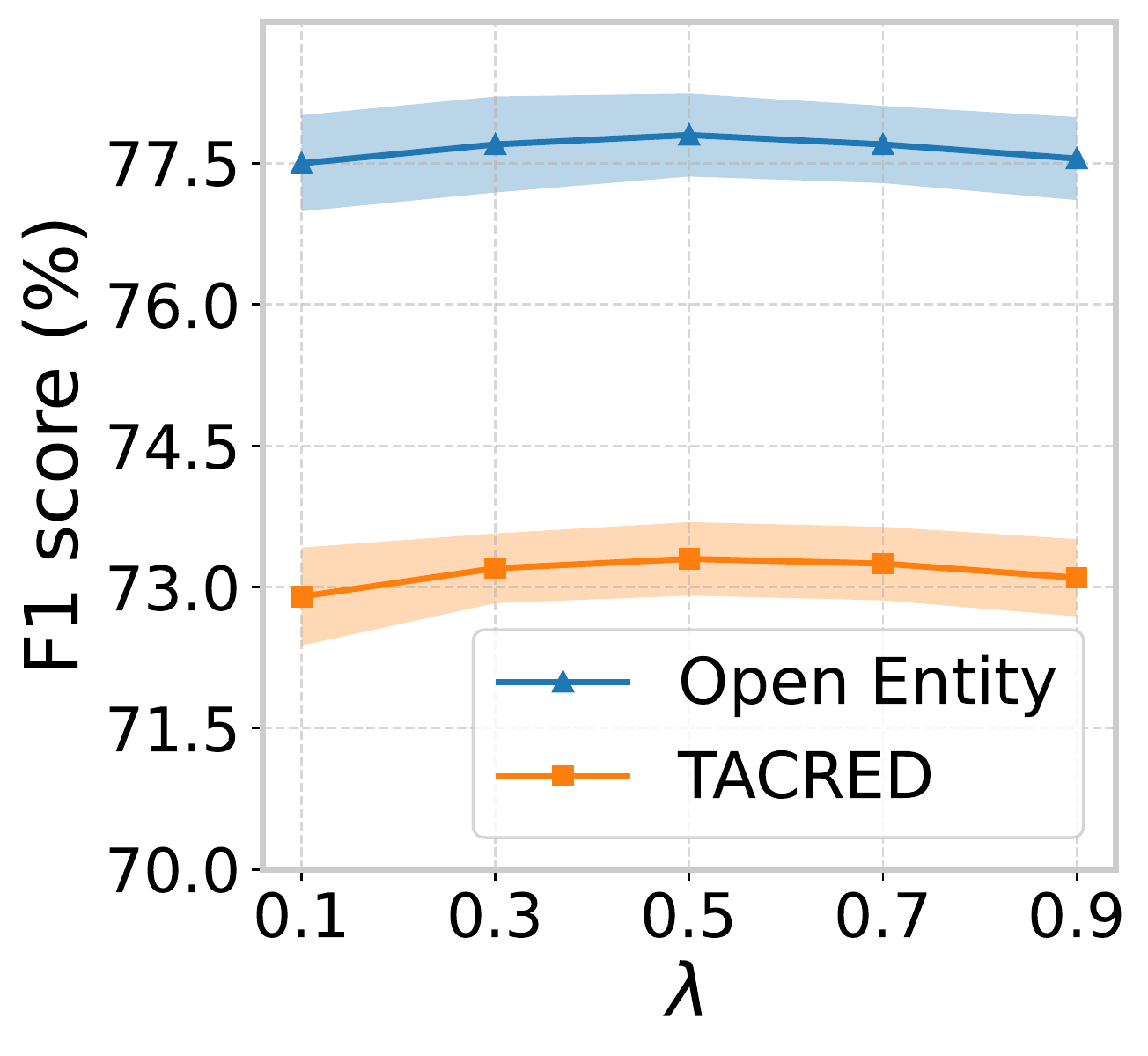}
\end{minipage}
\begin{minipage}[t]{0.48\linewidth}
    \includegraphics[width = 1\linewidth]{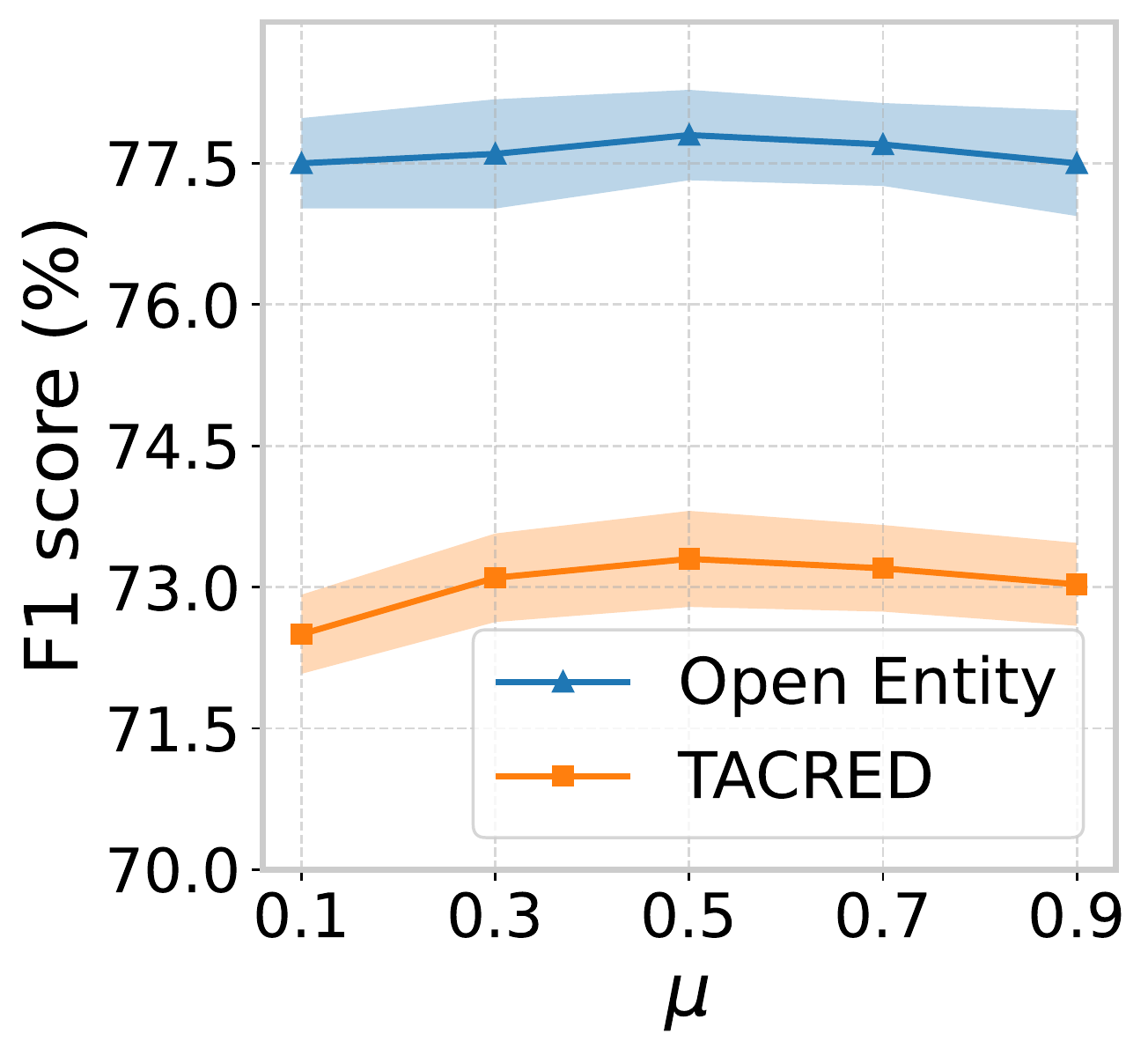}
\end{minipage}
\end{tabular}
\vspace{-.75em}
\caption{Hyper-parameter efficiency of $\lambda$ and $\mu$ over Open Entity and TACRED.}
\label{fig:hyper-parameter1}
\end{figure}

\begin{figure}[t]
\centering
\includegraphics[width=0.8\columnwidth]{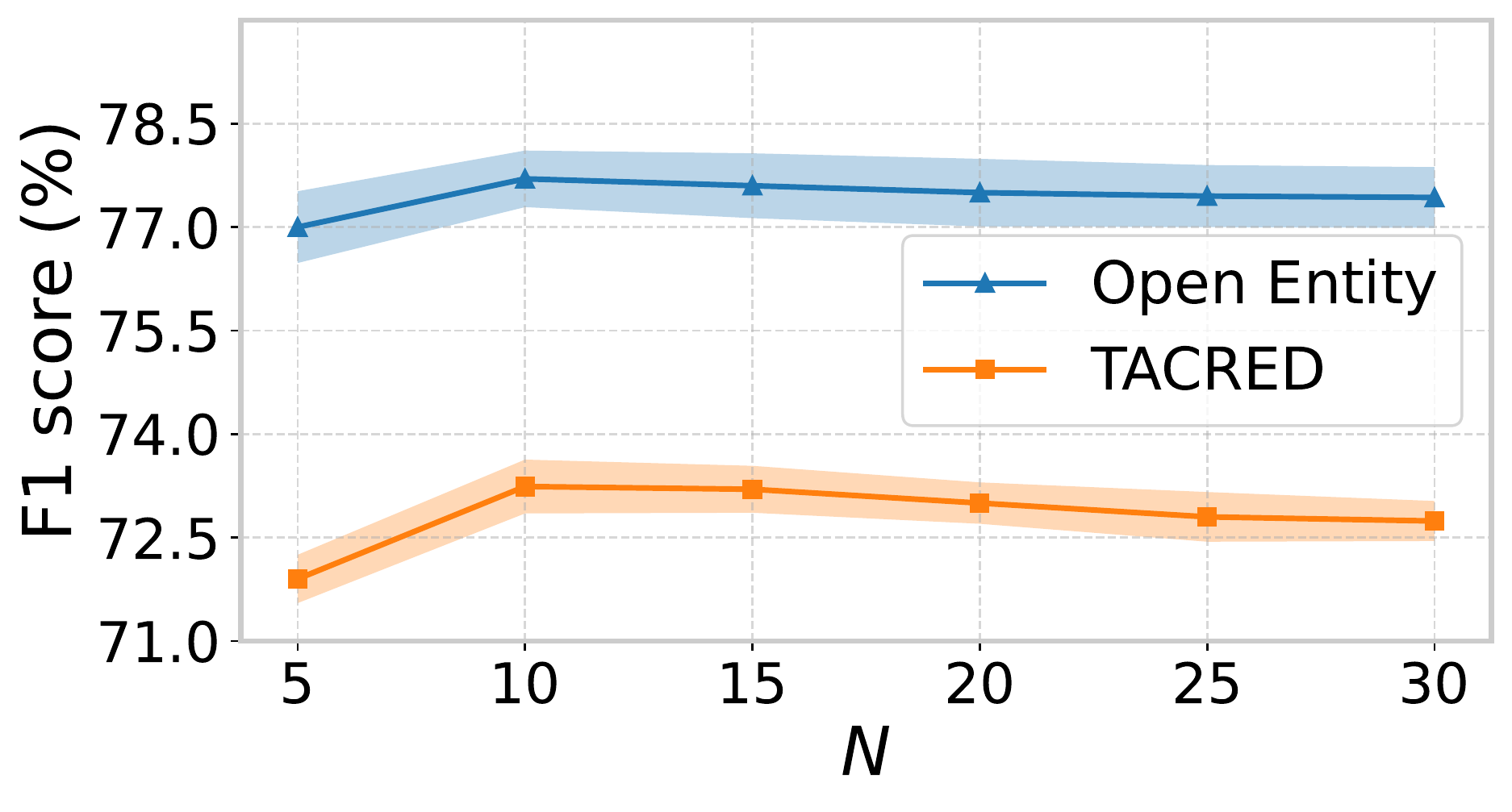}
\vspace{-.75em}
\caption{Hyper-parameter efficiency of $N$ over Open Entity and TACRED.}
\label{fig:hyper-parameter2}
\end{figure}

For the zero-shot learning, we directly use the vanilla MLM to generate the result at the \texttt{[MASK]} position. Formally, given an input sequence $X$, the label set $\mathcal{Y}$, and the verbalizer $\mathcal{M}$, the prediction can be calculated as:
\begin{equation*}
    \resizebox{\hsize}{!}{%
    $\hat{y} = \text{argmax}_{y\in\mathcal{Y}}(\text{Pr}_{\mathcal{F}}(\texttt{[MASK]} = \mathcal{M}(y)|X))$
    }
\end{equation*}
where $\mathcal{F}$ denotes the PLM, $\text{Pr}(\cdot)$ denotes the probability distribution generated by the MLM head of $\mathcal{F}$. For the regression task (i.e. SST-B), we directly use the probability value to represent the real-valued result.

For the few-shot learning, we follow~\citet{Gao2021Making} to construct few-shot  training/developing set, which consists only 16 examples for each label. For example, we have 32 examples for training and developing set for SST-2 dataset. During the training, the batch size is 4, the learning rate is set as 1e-5 with a warm up rate 0.1. We leverage the cross-entropy loss to train for 64 epochs, and evaluate the model over the whole testing set.

\noindent\textbf{Fine-tuning.}
For the full-data fine-tuning (FT-Full), we follow the standard supervised training settings (which can also be found in LM-BFF.





\begin{table}
\centering
\resizebox{\linewidth}{!}{
\begin{small}
\begin{tabular}{lc | cccc}
\toprule
\bf $k$ &\bf \#Entities &\bf Open Entity &\bf  TACRED & \bf FewRel & Avg. \\
\midrule
1 & 6 & 77.1 & 72.9 & 86.9 & 79.0 \\
2 & 15 & \textbf{77.8} & \textbf{73.3} & 87.5 & \textbf{79.5} \\
3 & 32 & 77.6 & 71.7 & \textbf{88.7} & 79.3 \\
4 & 62 & 75.8 & 69.9 & 85.4 & 77.0 \\
\bottomrule
\end{tabular}
\end{small}
}
\caption{The effectiveness (F1 value \%) of the $k$-hop sub-graph for~\emph{knowledge prompting}.}
\label{tab:knowledge-subgraph}
\end{table}

\subsection{Hyper-parameter Analysis}
\label{app:ha}

We also conduct a hyper-parameter sensitivity study on Open Entity and TACRED tasks.
Specifically, 
we study 
three key 
hyper-parameters: 
the balancing coefficients $\lambda$ and $\mu$ for PRI and MPM tasks in Eq.~\ref{eqn:total-loss}, 
and 
the number of negative entities 
$N$ in the MPM task. 
We vary one 
hyper-parameter with others fixed. 
From Figure~\ref{fig:hyper-parameter1}, 
as $\lambda$ ($\mu$) increases,
the performance of
{\model}\ first increases and then drops, and it
can achieve 
the best result when $\lambda=0.5$ ($\mu = 0.5$).
For $N$
in Figure~\ref{fig:hyper-parameter2}, 
we observe that 
the model performance improves significantly at the beginning, and then decreases when $N>10$. 
It shows that too many negative entities may introduce noise to degrade the model performance.

Finally, 
we analyze the effectiveness of the $k$-hop sub-graph. 
Specifically,
we construct 1-hop, 2-hop, 3-hop and 4-hop sub-graphs for knowledge prompting. 
We conduct experiments on Open Entity, TACRED and FewRel task. 
Results in Table~\ref{tab:knowledge-subgraph} suggest that we can achieve the best performance when $k=2$. 
We also find that the overall model performance decreases when $k>2$.
This may be explained by the introduction of 
many redundant and irrelevant entities.

\end{document}